\documentclass[runningheads]{llncs}

\usepackage{graphicx}
\usepackage{adjustbox}

\usepackage{amsmath}
\usepackage{amssymb}
\usepackage{color}
\usepackage{mathtools}
\usepackage{bm}
\usepackage{multirow}

\newcommand{\symspd}[1]{\ensuremath{\mathcal{P}(#1)}}

\usepackage{lineno}

\usepackage{pgfplots}
\pgfplotsset{compat=newest}

\newcommand{\NN}{\mathrm{NN}}
\newcommand{\NNp}{\ensuremath\phi_\NN}
\newcommand{\NNd}{\ensuremath\mathbf{d}_\NN}
\newcommand{\vx}{\mathbf{x}}

\newcommand{\tD}{\mathbf{D}}
\newcommand{\vd}{\mathbf{d}}
\newcommand{\tP}{\mathbf{P}}
\newcommand{\Resm}{R_\mathrm{m}}
\newcommand{\alpham}{\alpha_\mathrm{m}}
\newcommand{\vtheta}{\bm{\theta}}
\newcommand{\dmax}{d_\mathrm{max}}
\newcommand{\R}{\ensuremath{\mathbb{R}}}
\newcommand{\surf}{\ensuremath{\mathcal{S}}}
\newcommand{\diff}[1]{\ensuremath{\operatorname{d}\!{#1}}}
\newcommand{\RMSE}{\mathrm{RMSE}}
\newcommand{\RMSEV}{\RMSE_\mathrm{T}}
\newcommand{\RMSES}{\mathrm{RMSE}_{\mathcal{S}}}

\begin{document}
\def\tens  #1{\mbox{\boldmath{\scriptsize{$#1$}}}{}}
\def\ten   #1{\mbox{\boldmath $#1$}{}}

\title{Learning atrial fiber orientations and conductivity tensors
       from intracardiac maps using physics-informed neural networks%
       }

\titlerunning{Learning conductivity tensor from EAMs using PINNs}

\author{Thomas Grandits\inst{1,2} \and
        Simone Pezzuto\inst{3}\orcidID{0000-0002-7432-0424} 
        \and
        Francisco Sahli Costabal\inst{4,5,6} \and
        Paris Perdikaris\inst{7} \and
        Thomas Pock\inst{1,2}\orcidID{0000-0001-6120-1058} 
        \and
        Gernot Plank\inst{2,8}\orcidID{0000-0002-7380-6908} 
        \and
        Rolf Krause\inst{3}\orcidID{0000-0001-5408-5271}}

\authorrunning{Grandits et al.}

\institute{Institute of Computer Graphics and Vision, TU Graz,
Graz, Austria
\email{\{thomas.grandits,pock\}@icg.tugraz.at} \and
BioTechMed-Graz, Graz, Austria \and
Center for Computational Medicine in Cardiology,
Institute of Computational Science,
Universit\`a della Svizzera italiana,
Lugano, Switzerland
\email{\{simone.pezzuto,rolf.krause\}@usi.ch} \and
Department of Mechanical and Metallurgical Engineering,
School of Engineering, Pontificia Universidad Cat\'olica de Chile,
Santiago, Chile \and
Institute for Biological and Medical Engineering, Schools of Engineering,
Medicine and Biological Sciences, Pontificia Universidad Cat\'olica de Chile,
Santiago, Chile
\email{fsc@ing.puc.cl} \and
Millennium Nucleus for Cardiovascular Magnetic Resonance \and
Department of Mechanical Engineering and Applied Mechanics
University of Pennsylvania, Philadelphia, Pennsylvania, USA
\email{pgp@seas.upenn.edu} \and
Gottfried Schatz Research Center - Division of Biophysics,
Medical University of Graz, Graz, Austria
\email{gernot.plank@medunigraz.at}
}

\maketitle

\begin{abstract}

Electroanatomical maps are a key tool in the diagnosis and treatment of atrial fibrillation. 
Current approaches focus on the activation times recorded. 
However, more information can be extracted from the available data. 
The fibers in cardiac tissue conduct the electrical wave faster, and their direction could be inferred from activation times. 
In this work, we employ a recently developed approach, called physics informed neural networks, to learn the fiber orientations from electroanatomical maps, taking into account the physics of the electrical wave propagation. 
In particular, we train the neural network to weakly satisfy the anisotropic eikonal equation and to predict the measured activation times. 
We use a local basis for the anisotropic conductivity tensor, which encodes the fiber orientation. 
The methodology is tested both in a synthetic example and for patient data.
Our approach shows good agreement in both cases, with an RMSE of $2.2$ms on the in-silico data and outperforming a state of the art method on the patient data. 
The results show a first step towards learning the fiber orientations from electroanatomical maps with physics-informed neural networks.


\end{abstract}



\section{Introduction}

The fiber structure of the cardiac tissue has a prominent role on its function.
From an electrophysiological viewpoint, electrical conduction along the fiber
direction is generally higher than the cross-fiber
direction~\cite{clerc_fibers_1976}.  The overall propagation of the action
potential is therefore anisotropic, with preferential pathways in the
activation. In the atria, fiber orientations and the resulting conductivity are
however known only with large uncertainty, rendering the construction of
accurate computational models more difficult~\cite{roney_fibers_2020}.

Electroanatomical mapping, a keystone diagnostic tool in cardiac
electrophysiology studies, can provide high-density maps of the local electric
activation. 
Since conduction properties of the tissue and the activation times
are physiologically correlated, electroanatomical maps (EAMs) can be potentially
the basis for a parameter identification procedure of the
former~\cite{maagh_eam_2017}.
Subsequently, local fiber orientations may be
extrapolated from the fastest conductivity direction.

The problem of identifying the conductivity of the tissue from point-wise
recordings of the activation time has been already addressed by others.  Local
approaches estimate the front velocity at every point of the domain only by
using the local geometrical and electrical information, that is the location
and the activation time of neighboring points~\cite{verma_cv_2018}. Local fiber
direction can be estimated from multiple pacing sequences by comparing the
velocity in the propagation directions~\cite{roney_fibers_2019}.  Similarly,
the EAM may be interpolated into a smooth activation map, e.g., with linear or
radial basis functions~\cite{coveney_gp_2020}.
Local methods are general and do not account for the physics:
there is no guarantee that activation and conduction will satisfy a given
model.
Moreover, they may fail at
those locations
where activation is not differentiable, such as collision lines and
breakthroughs. 
Model-based approaches are not novel
either~\cite{grandits_inverse_2020,barone_cardiac_2020}.  In
these methods, the mismatch between observed and simulated (by the model)
activation is minimized by optimizing the local conduction velocity of the
tissue. 

In the presented work, we solve the problem of learning the anisotropic structure of the
conductivity tensor from electroanatomical maps by imposing to the
identification problem a physiological constraint, encoded in the anisotropic
eikonal equation.  For this purpose, we extend our recent work into this
problem~\cite{sahli_costabal_eikonalnet_2020,grandits_piemap_2021} by combining
Physics-informed Neural Networks (PINNs) and the anisotropic eikonal equation.
In the presented approach, a set of neural networks representing the conductivity tensor
and the activation times are fitted to the data by weakly imposing the eikonal
model through a penalization term.  PINNs are particularly suitable for
recovering complex functions from sparse and scarce data, as in the present
case~\cite{raissi_pinn_2019}. Moreover, the weak imposition of the model does
not require its explicit solution.  
In contrast to prior learning methods, aiming at learning Finite Element Method (FEM) solutions~\cite{takeuchi_neural_1994}, PINNs learn a single (albeit complex) continuous function to estimate the solution on the whole domain.
Therefore, PINNs can model functions on domains irrespective of a FE discretization, i.e.~mesh.
The function can in this setting also be evaluated outside the domain, where it is not required to weakly satisfy the model partial differential equation (PDE).

We can also omit boundary conditions, which for the eikonal model are the sites of early activation, usually unknown.
The proposed method, fully implemented in TensorFlow\footnote{\url{https://www.tensorflow.org/}},
performed  well with synthetic data, being able to recover ground-truth fiber
orientations in limited regions, just with a single activation sequence. 
Additionally, we applied the method to EAMs from a patient who underwent a
mapping procedure prior ablation, with promising results.
Both models were compared to our previous classical optimization approach \cite{grandits_piemap_2021} and performed comparably in all cases.

This manuscript is organized as follows: In Section~\ref{sec:methods} we
formulate the problem of estimating conductivities as a physics-informed neural
network, and we introduce its numerical solution. In Section~\ref{sec:numexp},
we show two numerical experiments to test the accuracy of the method. We end
this manuscript with a discussion and future directions in
Section~\ref{sec:discussion}.


\section{Methods}

\label{sec:methods}

\subsection{Anisotropic eikonal model for cardiac activation}

The cardiac tissue is electrically active, in the sense that a sufficiently
strong stimulus can trigger the propagation of a travelling wave, called action
potential.  The time of first arrival of the action potential, usually set as
the point of crossing of a threshold potential, is called activation time.
Herein, we denote as $\phi\colon\Omega\to\mathbb{R}$ the activation map in a
embedding domain $\Omega \subset \R^3$ (e.g., a box containing the atria), that is
$\phi(\vx)$ is the activation time at $\vx\in\Omega$. 
Neglecting curvature effects of the wave on the propagation speed~\cite{colli_eikonal_1990},
the activation map may be well described by the
anisotropic eikonal equation, which reads as follows:
\begin{equation}\label{eq:aniso_eikonal}
\sqrt{\tD(\vx) \nabla\phi(\vx)\cdot\nabla\phi(\vx)} = 1,
\quad\vx\in\Omega,
\end{equation}
where $\tD\in\mathcal{C}(\bar\Omega;\symspd{3})$ is a continuous tensor
field from $\bar\Omega$ to $\symspd{3}$, the set of $3\times 3$ symmetric
positive-definite real matrices. 
The eikonal equation is not explicitly solved, hence there is no need to enforce boundary conditions.
We rather consider the model residual:
\begin{equation}
\Resm[\phi](\vx) \coloneqq \sqrt{\max\bigl\{ \tD(\vx) \nabla\phi(\vx)\cdot
\nabla\phi(\vx),\varepsilon\bigr\} } - 1,
\label{eq:aniso_eikonal_residual}
\end{equation}
for a sufficiently small $\varepsilon > 0$ to avoid infeasible gradients,
as a metric of point-wise model discrepancy for a given pair of activation
$\phi$ and conductivity tensor $\tD$.


\subsection{Representation of the conductivity tensor}
\label{sec:repr}
\begin{figure}[htb]
    \centering
    \includegraphics[width=.75\textwidth]{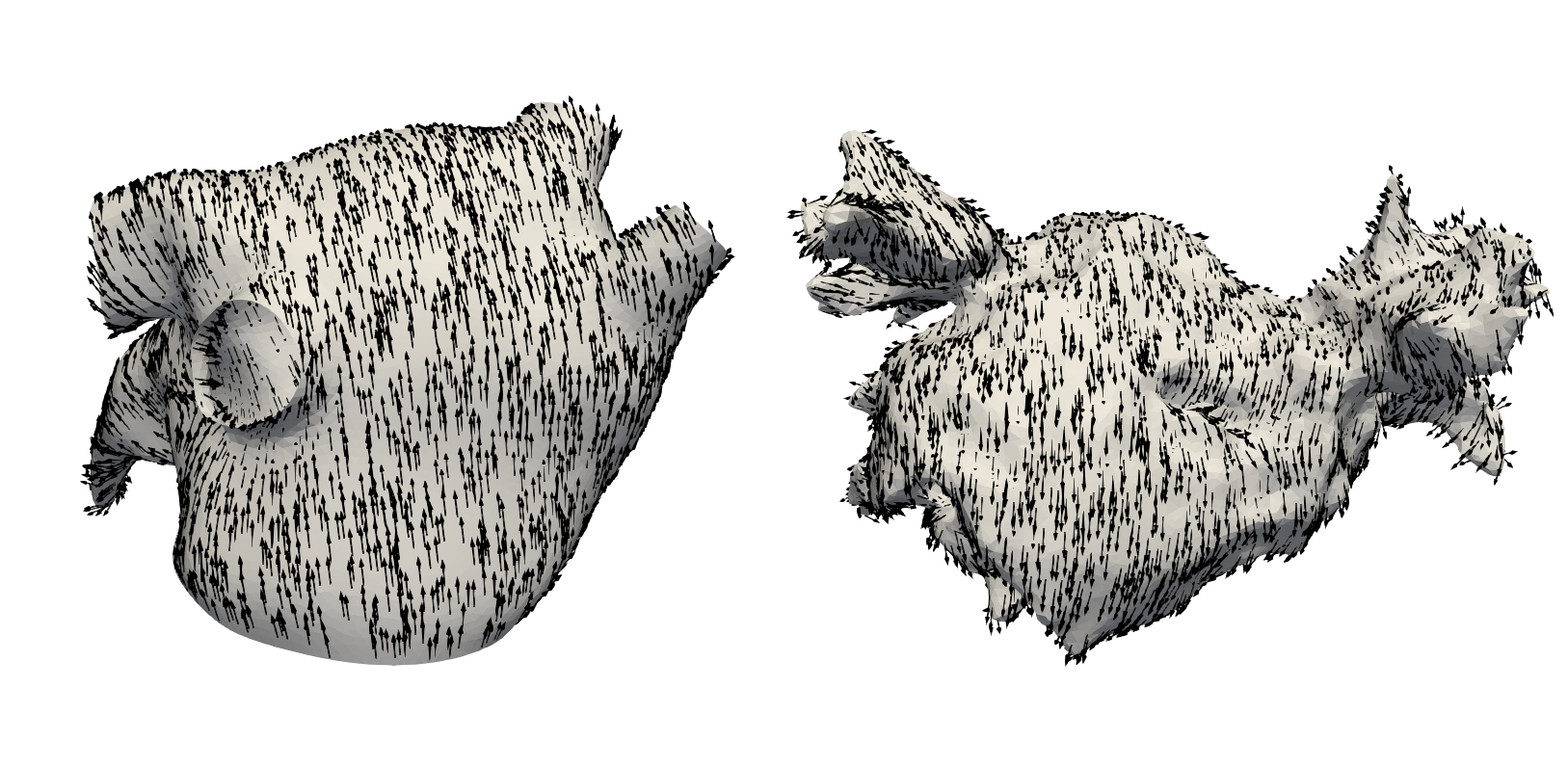}
    \caption{Smooth generated manifold bases using the vector heat method for both considered models.
    These bases are used to provide a smooth 2D map across the manifold and are a useful foundation for computing TV in 2D.}
    \label{fig:smooth_bases}
\end{figure}

The conductivity tensor $\tD$ shall be defined through a parameter vector
in a way that ensures its symmetry, positive-definiteness, and zero velocity orthogonal to the atrial
surface $\surf\subset\Omega$ in all cases. 
Analogous to our previous method in \cite{grandits_piemap_2021}, we therefore consider the parameter vector
$\vd(\vx) = [d_1(\vx), d_2(\vx), d_3(\vx)]^\top$ and define $\tD$ through $\vd$ as follows:
\begin{equation}\label{eq:expD}
\tD(\vx) \coloneqq e^{\tP(\vx) \tD_2(\vd(\vx)) \tP(\vx)^\top},
\qquad \tD_2(\vd(\vx)) \coloneqq
\begin{bmatrix} d_1(\vx) & d_2(\vx) \\ d_2(\vx) & d_3(\vx) \end{bmatrix}
\end{equation}
where $\tP(\vx)\in\mathbb{R}^{3\times 2}$ is a matrix whose columns contain two
orthonormal vectors in the tangent plane at $\vx\in\surf$, computed using
the vector heat method~\cite{sharp_vectorheat_2019}. 
The smooth tangent bases generated by the vector heat method for both the in-silico model, as well as the in-vivo measured EAM can be seen in Fig.~\ref{fig:smooth_bases}.

Please note that we
consider the matrix exponential in Eq.~\eqref{eq:expD} rather than
component-wise exponentiation.  This choice corresponds to the Log-Euclidean
metric of the space $\symspd{3}$~\cite{arsigny_geometric_2007}.

The fiber direction is defined as the direction of fastest propagation, that is
the eigenvector associated to the largest eigenvalue of $\tD$. By the
definition of matrix exponential, such direction is also the maximum
eigenvector of $\tP \tD_2 \tP^\top$.


\subsection{Physics-informed neural network}

In the considered experiments, we are given a set of points, each composed by a location $\vx_i$ and a
recorded activation time $\hat\phi_i: \Gamma_\surf \to \R$  for $\Gamma_\surf \subset \surf$ 
representing the EAM locations and timings of the recordings.  
The objective is therefore to identify a conductivity
tensor field $\tD$ such that the corresponding activation map $\phi$, as
resulting from Eq.~\eqref{eq:aniso_eikonal}, will closely reproduce the
observed data.
The tensor $\tD$ can then be reconstructed by means of using $\vd$.

\begin{figure}[htb]
    \centering
    \adjincludegraphics[Clip={{.00 \width} {.00 \height} {.00 \width} {.0 \height}}, width=.75\textwidth]{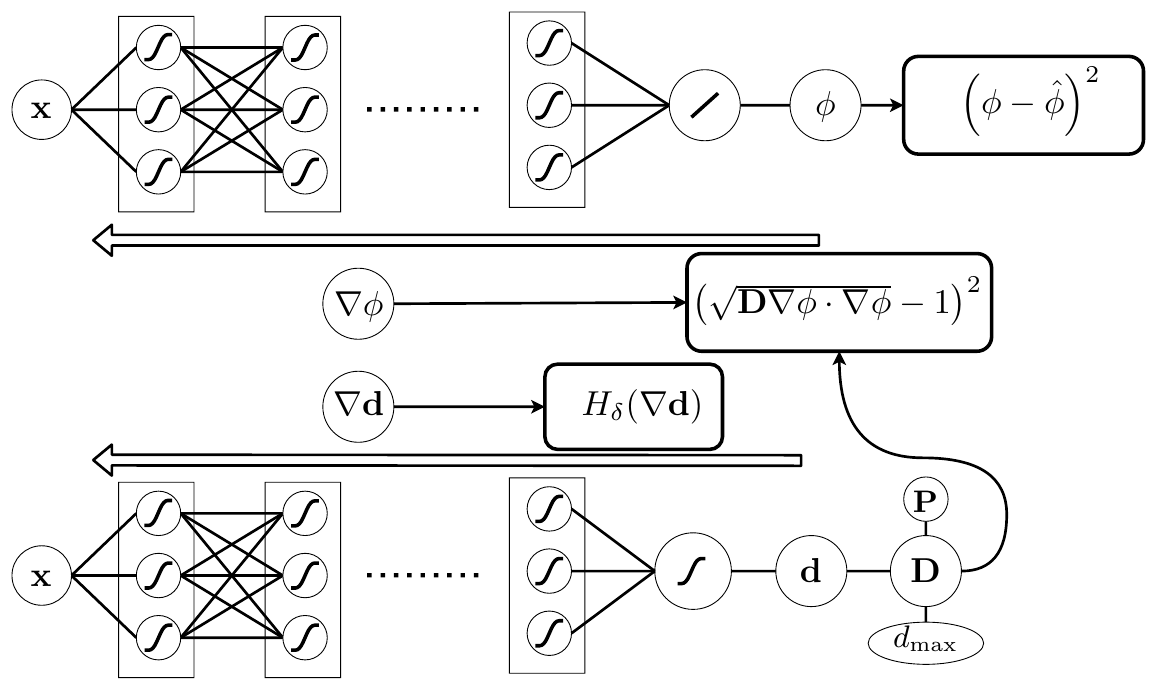}
    \caption{Structural view of the proposed PINN architecture, containing the two NNs $\NNp$ and $\NNd$.
    Nodes containing a curve indicate a tanh activation function.
    The final layer of $\NNp$ is a linear layer.
    $\tD$ is computed using~\eqref{eq:expD}.
    $\nabla \phi$ and $\nabla \vd$ can be obtained by means of backpropagation (reverse arrows).
    The bold rectangular boxes show the three major loss terms from~\eqref{eq:loss}: 
    Data fidelity, eikonal (PDE) and TV loss (top to bottom).}
    \label{fig:neural_network}
\end{figure}

For this purpose, we approximate both the activation map $\phi(\vx)$ and the conductivity
vector $\vd(\vx)$ with a feed-forward neural network
$\NN_{n,m,\vtheta} \colon \R^n\to \R^m$ with
$n$ inputs to $m$ outputs and characterized by
a vector $\vtheta$ containing weights and biases, as was initially promoted in \cite{raissi_pinn_2019}.
The used architecture of the networks is shown in Fig.~\ref{fig:neural_network}.
Specifically, we have $\phi(\vx) \approx \NNp(\vx, \vtheta_\phi) = \NN_{3,1,\vtheta_\phi}(\vx)$ and $\vd(\vx)  \approx \NNd(\vx, \vtheta_{\vd}) = \dmax\cdot\tanh\bigl( \NN_{3,3,\vtheta_{\vd}}(\vx) \bigr)$.
where $\tanh$ is meant component-wise, and $\dmax$ is an upper limit for the
components of $\NNd$, meant to avoid over- and underflows in the numerical calculations.
This construction of $\phi$ and $\vd$ enables us to use standard machine learning methods and frameworks to efficiently calculate the gradients $\nabla \phi$ and $\nabla \vd$, used in the chosen PDE model~\eqref{eq:aniso_eikonal_residual} and inverse regularization.
The usage of these gradients in the optimization necessitates at least second order smooth activation functions in the neurons, achieved by the use of $\tanh$ functions.

Similar to the original PINN algorithms in \cite{raissi_pinn_2019}, we define a loss function to train our model as the sum of a data fidelity,
a PDE model fidelity term and two regularization terms:
\begin{equation}
\begin{split}
    \mathcal{L}(\vtheta_\phi, \vtheta_{\vd}) &\coloneqq
    \int_{\Gamma_\surf} \bigl( \NNp(\vx) - \hat{\phi}(\vx) \bigr)^2 \diff{\vx}
  + \alpham \int_\surf \bigl( \Resm[\NNp](\vx)\bigr)^2 \diff{\vx} \\
 &+ \alpha_\theta \bigl( \|\vtheta_\phi \|^2 + \|\vtheta_\vd \|^2 \bigr)
 +  \alpha_{\vd} \int_\surf H_{\delta}\bigl( \nabla \NNd(\vx) \bigr) \diff{\vx},
\label{eq:loss}
\end{split}
\end{equation}
for the three weighting parameters $\alpham, \alpha_{\vd}, \alpha_\theta$.
Regularization is both applied to the weights of the networks as well as on the inverse parameter estimation.
The latter regularization term is an Huber-type, approximated Total Variation regularization
for the conductivity vector parameters $\vd$. Specifically,
\begin{equation}
    H_{\delta}(\mathbf{x}) = 
    \begin{cases}
     \frac{1}{2 \delta} 
     \|\mathbf{x}\|^2, & \mbox{if $\|\vx\| \le \delta$,} \\
     \|\mathbf{x}\| - \frac{1}{2} \delta, & \mbox{otherwise}
    \end{cases}
\label{eq:huber}
\end{equation}
with $\delta = 5\cdot 10^{-2}$ for our experiments. 
Note that by construction, $\tD$ has zero velocity in directions normal to the manifold (see Sec.~\ref{sec:repr}) and thus allows us to neglect the additional normal penalization used in \cite{sahli_costabal_eikonalnet_2020}.

\subsection{Numerical implementation}
\label{sec:num}
The domain $\surf$ is discretized using a triangular mesh, usually obtained directly from the mapping system, along with point-wise evaluations of the activation times, that is $\Gamma_\surf = \{ \vx_1, \ldots, \vx_N \}$.  
In all experiments, the integrals were approximated using a point-wise evaluation for both domains: 
On the vertices for the approximation of $\surf$ and on the discrete measurements for $\Gamma_\surf$.

For the optimization, we experimentally selected the hyper-parameters as $\alpham = 10^4$ for the model atria and $\alpham = 10^3$ for the EAM.
The other two hyperparameters are the same for both experiments: $\alpha_\theta = 10^{-4}$, $\alpha_{\vd} = 10^{-3}$.
The two neural networks for $\phi$ and $\vd$ had 7 and 5 hidden fully connected layers respectively.
All hidden layers consisted of 20 neurons for $\NNp$ and 5 neurons for $\NNd$, with the weights being initialized using Xavier initialization. 
This choice of neural network architecture was inspired by the work in~\cite{sahli_costabal_eikonalnet_2020}.
We opted for adding a regression layer to $\NNp$, since this allows us to model arbitrary ranges of $\phi$.
Optimization is performed by first using the ADAM \cite{kingma2014adam} optimizer for $10^4$ epochs with a learning rate of $10^{-3}$, followed by a L-BFGS optimization \cite{byrd1995limited} until convergence to a local minimum is achieved.
Each experiment took no longer than 1.5~hours on a desktop machine with an Intel Core i7-5820K CPU with 6 cores of each 3.30GHz, 32GB of working memory and a NVidia RTX 2080 GPU.


\section{Numerical experiments}
\label{sec:numexp}


Herein, we consider two experiments: a synthetic example, with ground-truth on a realistic anatomy of the left atrium, and an example with patient-specific geometry and data.  
The first example is optimized and tested against different levels of i.i.d.\ normal noise: $\tilde{\phi}(\vx) = \hat{\phi}(\vx) + \mathcal{N}(0, \sigma_\mathcal{N})$.  
We measure the performance of the synthetic model in terms of the root-mean-square error (RMSE) over the whole surface here denoted as $\RMSES$.
Errors directly on the measurement points, employed in the optimization, are used to compute $\RMSE_O$.
In the patient specific example, we randomly split $\Gamma_\surf$ into $\Gamma_O$, used for optimization/training, and $\Gamma_T$, for testing.
%

The results of our method on the in-silico model atria, and a comparison to PIEMAP \cite{grandits_piemap_2021}, are presented in Tab.~\ref{tab:comparison_tab_model}. 
Both methods are comparable in terms of RMSE, with both methods achieving less than $5$ms of RMSE for all levels of noise.
Additionally, we tested the presented PINN on an EAM, achieving a RMSE of the activation times on the test set of $\RMSEV \approx 5.59$\,ms.
The RMSE on the measurements used in the optimization was only slightly lower at $\RMSE_O \approx 4.82$\,ms, indicating that $\alpham$ was chosen in a proper range to avoid overfitting to the data.
In the patient-specific test (not shown in the table), our method was able to outperform PIEMAP, which reported $\RMSEV \approx 6.89$ms and $\RMSE_O \approx 1.18$ms on test and optimization/training set respectively, showing a slight overfit to the data used in the optimization. 

\bgroup
\def\arraystretch{1.2}
\setlength{\tabcolsep}{.4em}
\footnotesize
\begin{table}[htb]
    \centering
    \caption{Evaluation of the presented PINN approach compared to PIEMAP \cite{grandits_piemap_2021} for different levels of noise (given in standard deviation and signal-to-noise ratio) on the in-silico model atria.
            The result of the noiseless scenario ($\infty$ dB) is visualized in Fig.~\ref{fig:results} on the left side.}
    \begin{tabular}{cc|c|c|c|}
    \cline{3-4}
         &   & $\RMSES/\RMSE_O$ PINN  & $\RMSES/\RMSE_O$ PIEMAP \\ \hline
         \multicolumn{1}{|l|}{\multirow{4}{*}{\rotatebox[origin=c]{90}{$\sigma_{\mathcal{N}}$/PSNR}}} & 0\,ms/$\infty$ dB & 2.20/1.38 & 1.04/0.83 \\ \cline{2-4}
    \multicolumn{1}{|l|}{}       & 0.1\,ms/64.1 dB & 4.28/2.08 & 1.02/0.83 \\ \cline{2-4}
     \multicolumn{1}{|l|}{}       & 1\,ms/43.9 dB & 3.32/1.39 & 1.09/0.83 \\ \cline{2-4}
    \multicolumn{1}{|l|}{}       & 5\,ms/29.9 dB & 3.76/1.85 & 1.90/0.84 \\ \hline
    \end{tabular}
    \label{tab:comparison_tab_model}
\end{table}
\egroup

\begin{figure}[htb]
    \centering
    \begin{tikzpicture}
        \node (model_atria)
        {\adjincludegraphics[Clip={{.125 \width} {.025 \height} {.17 \width} {.025 \height}}, width=.41\textwidth]{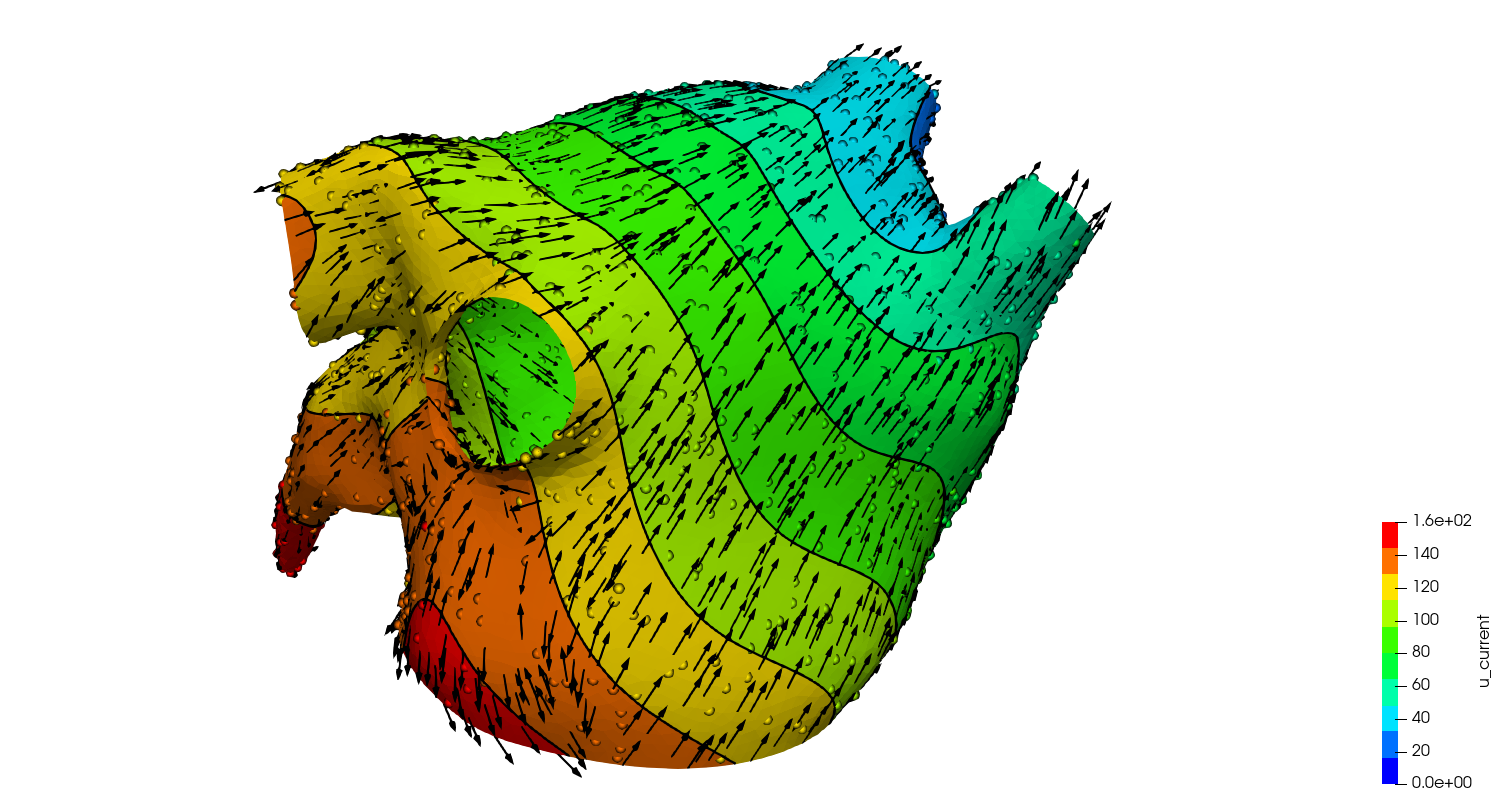}};
        \node[anchor=west] (eam) at (model_atria.east)
        {\adjincludegraphics[Clip={{.125 \width} {.07 \height} {.17 \width} {.075 \height}}, width=.41\textwidth]{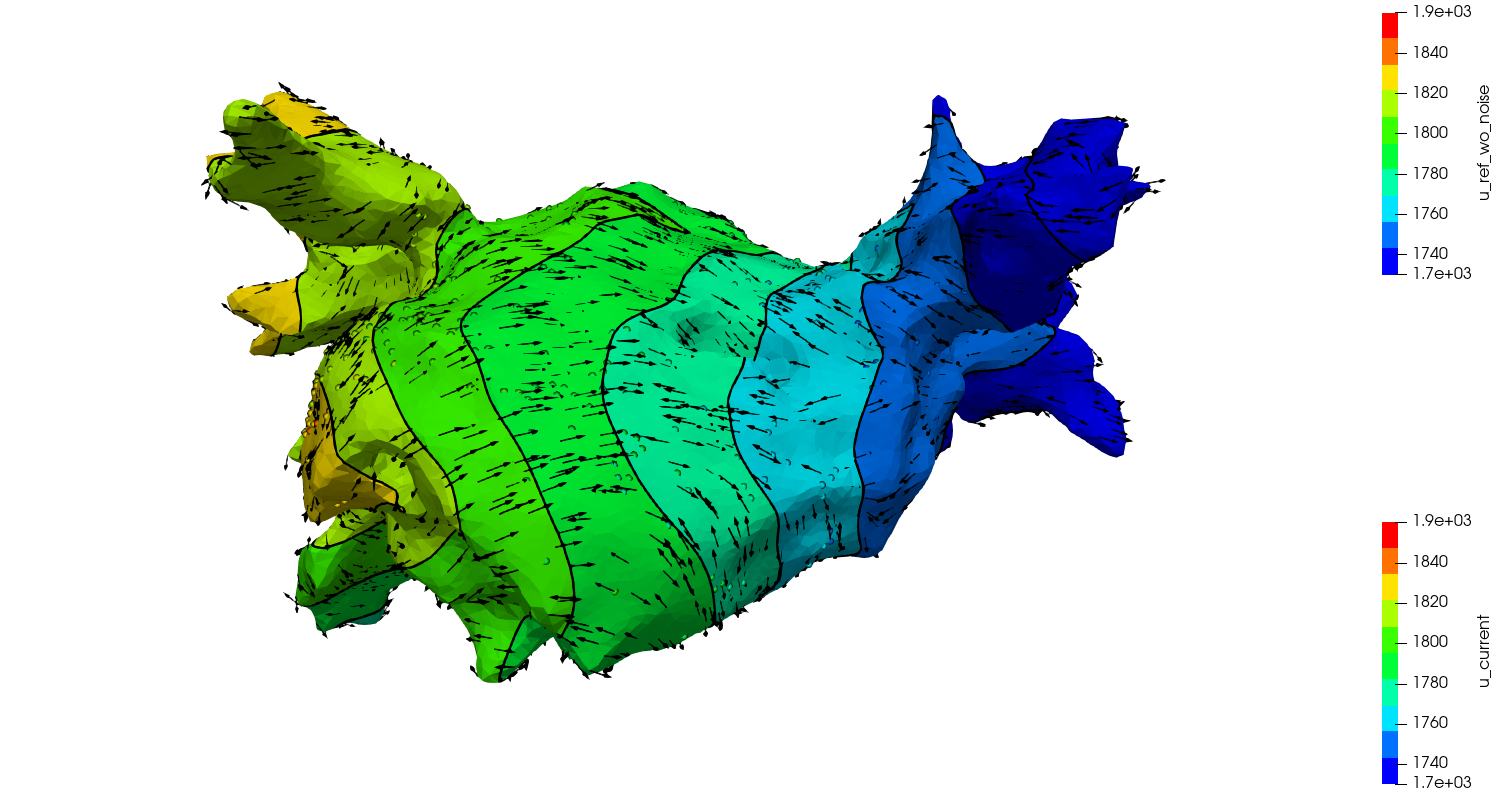}};
        \node[anchor=north, yshift=.25cm] (model_atria_cbar) at (model_atria.south)
        {   
            \pgfplotscolorbardrawstandalone[ 
            colormap/bluered,
            colorbar horizontal,
            colorbar sampled,
            point meta min=0,
            point meta max=160,
            colorbar style={
                samples=11,
                title=Activation Times $\phi$ [ms],
                title style={
                    font=\small,
                    yshift=-1.5mm
                },
                height=.15cm,
                width=.35\linewidth,
                xtick={0,32,...,160}},
            ticklabel style={
                font=\small}]
        };
        \node[anchor=west] (eam_cbar) at (model_atria_cbar.east)
        {   
            \pgfplotscolorbardrawstandalone[ 
            colormap/bluered,
            colorbar horizontal,
            colorbar sampled,
            point meta min=0,
            point meta max=130,
            colorbar style={
                samples=11,
                title=Activation Times $\phi$ [ms],
                title style={
                    font=\small,
                    yshift=-1.5mm
                },
                height=.15cm,
                width=.35\linewidth,
                xtick={0,26,...,130}},
            ticklabel style={
                font=\small}]
        };
    \end{tikzpicture}
    \caption{Results of the PINN method on an in-silico (left) and an in-vivo (right) EAM model with the overlayed measurements as points and fibers as arrows.
    The underlying contour lines and colors on the mesh itself represents the activation of the PINN, sampled at each vertex.
    }
    \label{fig:results}
\end{figure}

Fig.~\ref{fig:results} shows the qualitative results of using this method on the two chosen models (the model atria in the noise-less case).
We can nicely fit the activation encountered at the surface and create an eikonal-like activation.
The initiation sites are automatically deduced by the PINN algorithm with only soft eikonal and data constraints.
The smooth basis generated with the vector heat method, together with the TV regularization give us a smooth, fiber field.

\section{Discussion and outlook}
\label{sec:discussion}
In this work, we present a novel approach to the recovery of fiber orientations from activation time measurements. 
We train a neural network that aims to approximate the data while also satisfy the eikonal equation. 
In this way, we can identify the conductivity tensor that best explains the data. 
Our approach has some advantages compared to methods that exactly solve the forward problem of the eikonal equation. 
First, we only weakly enforce the solution of the PDE, which may be convenient when the model is not exactly satisfied. 
Here the eikonal equation is an approximation to the much complex process of cardiac electrophysiology. 
Second, we do not need to impose boundary conditions, or define the number of early activation sites, which are unknown beforehand. 
Third, our approach scales wells both with the number of measurements and the size of the mesh. 
We take advantage of mini-batch optimization strategies to utilize all the data while keeping the computational cost manageable. 
Finally, the implementation is compact and easily portable, taking advantage of the automatic differentiation capabilities of modern machine learning languages. 

Our work also has some limitations and disadvantages. 
First, the PINN optimization in~\eqref{eq:loss} is highly non-linear and exposes many local minima.
A direct use of a simple convex optimization algorithm is therefore often not advised as usually only very poor local minima are found.
ADAM can overcome some of these minima, but the final solution is very dependent on the initialization.
This could be mitigated by training multiple neural networks at a negligible cost, which at the same time can be used to quantify uncertainty \cite{sahli_costabal_eikonalnet_2020}. 
Future research could mitigate some of these problems by using different loss functionals, or alternative optimization algorithms.
Second, our method showed worse performance in terms of error when compared to PIEMAP in a synthetic example. 
However, this trend  was reversed when we used activation times coming from a patient. 
This indicates that further testing is necessary, with a larger dataset, to determine the accuracy of this method. 
Finally, our approach introduces multiple hyper-parameters that need to be tuned. In the future, we plan to use a more systematic approach to determine these parameters.

Overall, our work represent a first step to learning the fiber orientations from activation times in cardiac electrophysiology with physics-informed neural networks. 



\subsubsection*{Acknowledgments.}
This work was financially supported by the Theo Rossi di Montelera Foundation, the Metis Foundation Sergio Mantegazza, the Fidinam Foundation, the Horten Foundation and the CSCS--Swiss National Supercomputing Centre production grant s1074.
We'd also like to additionally acknowledge funding from the BioTechMed ILearnHeart project, the EU grant MedalCare 18HLT07, the grant FONDECYT-Postdoctorado 3190355, as well as the Swiss National Science Foundation for their support under grant 197041 ``Multilevel and Domain Decomposition Methods for Machine Learning''.

\bibliographystyle{splncs04}
\bibliography{fimh2020}

\end{document}